\pdfoutput=1

\documentclass[11pt]{article}

\usepackage[]{EACL2023}

\usepackage{times}
\usepackage{latexsym}

\usepackage[T1]{fontenc}

\usepackage[utf8]{inputenc}

\usepackage{microtype}

\usepackage{inconsolata}

\usepackage{graphicx}
\usepackage{xcolor}
\usepackage{CJKutf8}
\usepackage{tabularx}
\usepackage{booktabs}
\usepackage{makecell}
\usepackage{url}

\title{Cross-Lingual Question Answering over Knowledge Base\protect\\as Reading Comprehension}

\author{
    Chen Zhang$^{1}$, 
    Yuxuan Lai$^{3}$, 
    Yansong Feng$^{1,2}$\thanks{\quad Corresponding author.}, \\
    {\bf Xingyu Shen}$^{1}${\bf ,} 
    {\bf Haowei Du}$^{1}${\bf ,}
    {\bf Dongyan Zhao}$^{1,4,5}$ \\
    $^{1}$ Wangxuan Institute of Computer Technology, Peking University, China\\
    $^{2}$ The MOE Key Laboratory of Computational Linguistics, Peking University, China\\
    $^{3}$ Department of Computer Science, The Open University of China \\
    $^{4}$ National Key Laboratory of General Artificial Intelligence \\
    $^{5}$ Beijing Institute for General Artificial Intelligence \\
    {\tt \{zhangch,fengyansong,shenxy,zhaody\}@pku.edu.cn} \\
    {\tt duhaowei@stu.pku.edu.cn}~~
    {\tt laiyx@ochn.edu.cn} \\
}

\begin{document}
\maketitle
\begin{abstract}
Although many large-scale knowledge bases (KBs) claim to contain multilingual information, their support for many non-English languages is often incomplete. 
This incompleteness gives birth to the task of cross-lingual question answering over knowledge base (xKBQA), which aims to answer questions in languages different from that of the provided KB. 
One of the major challenges facing xKBQA is the high cost of data annotation, leading to limited resources available for further exploration.
Another challenge is mapping KB schemas and natural language expressions in the questions under cross-lingual settings. 
In this paper, we propose a novel approach for xKBQA in a reading comprehension paradigm.
We convert KB subgraphs into passages to narrow the gap between KB schemas and questions, which enables our model to benefit from recent advances in multilingual pre-trained language models (MPLMs) and cross-lingual machine reading comprehension (xMRC).  
Specifically, we use MPLMs, with considerable knowledge of cross-lingual mappings, for cross-lingual reading comprehension.
Existing high-quality xMRC datasets can be further utilized to finetune our model, greatly alleviating the data scarcity issue in xKBQA. 
Extensive experiments on two xKBQA datasets in 12 languages show that our approach outperforms various baselines and achieves strong few-shot and zero-shot performance. 
Our dataset and code are released for further research\footnote{\url{https://github.com/luciusssss/xkbqa-as-mrc}}.
\end{abstract}

\section{Introduction}
Large-scale knowledge bases (KBs) such as Freebase \cite{bollacker2008freebase} and DBpedia \cite{auer2007dbpedia} store huge amounts of structured knowledge. 
These KBs support a variety of natural language processing tasks, including question answering over knowledge base (KBQA), where models exploit the knowledge related to the questions and precisely identify the answers by reasoning through various KB relations.  
Although most large-scale KBs claim to contain multilingual information, they could not completely support non-English languages as expected. 
For example, Freebase has no translation for the KB relations/attributes in any non-English languages. 
More than half of the entities in Freebase have no Chinese translations, despite the fact that Chinese is the most spoken non-English language in the world. 
Therefore, these KBs could not directly support question answering in non-English languages, bringing up the problem of answering non-English questions over the KBs constructed in English.

\begin{figure}
\centering
\includegraphics[scale=0.3]{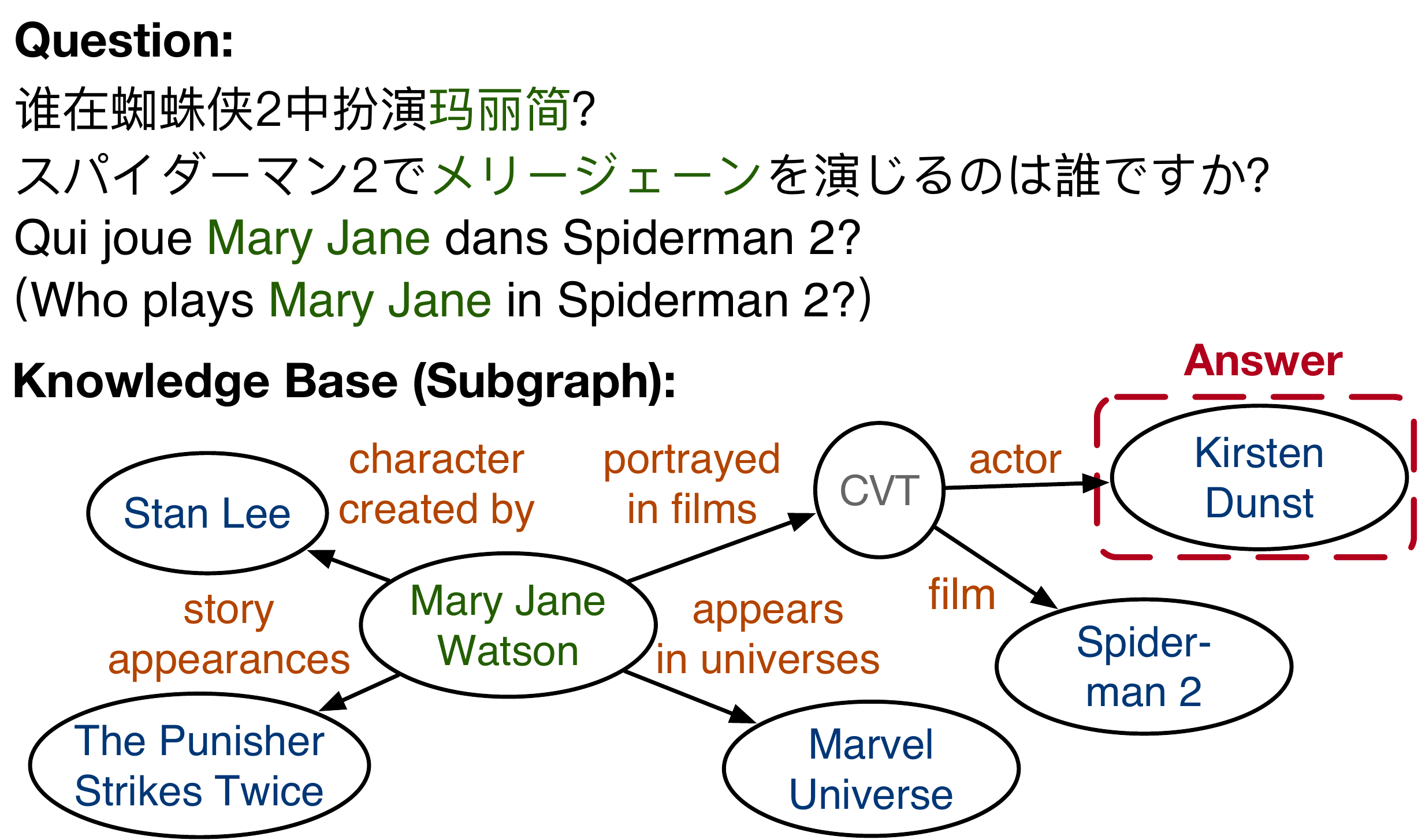}
\caption{An example of answering questions in non-English languages over an English knowledge base.}
\label{fig:xkbqa_example}
\end{figure}

In this work, we focus on cross-lingual KBQA (xKBQA), which aims to answer questions over a KB in another language.
Figure~\ref{fig:xkbqa_example} shows a KB subgraph and several factoid questions in non-English languages, which can be answered by a node in the KB subgraph.
Despite considerable progress in monolingual KBQA, xKBQA receives little attention. 
A significant challenge in xKBQA is the lack of large-scale xKBQA datasets. 
Such datasets are quite expensive to annotate since the annotators are expected to be multilingual and have background knowledge about KBs.
As a result, even the largest xKBQA dataset so far contains only a few hundred questions~\cite{ngomo20189th}. 
Another challenge is that, compared to other cross-lingual tasks, the expression difference between structured KB schemas and natural language questions further hinders the learning of cross-lingual mapping.

To address these challenges, we propose to convert the KB subgraphs into natural language texts and leverage the progress in cross-lingual machine reading comprehension (xMRC) to solve the xKBQA task.
Recently, there has been a series of large-scale xMRC datasets, such as MLQA \cite{lewis2020mlqa}, MKQA \cite{longpre-etal-2021-mkqa} and XQuAD \cite{artetxe2020cross}. 
Multilingual pre-trained language models (MPLMs), such as mBERT \cite{devlin2019bert} and XLM-R \cite{Conneau2020UnsupervisedCR}, achieve competitive performance on these xMRC benchmarks. 
As for xKBQA, by converting KB subgraphs into natural language texts, we narrow the gap between KB schemes and natural language expressions. 
We then utilize the PLM-based xMRC models finetuned on xMRC datasets to learn the cross-lingual mapping efficiently, even with limited xKBQA annotations.

Specifically, we first identify the topic entity from the given question, link it to the KB, and extract its $n$-order neighbors to construct a KB subgraph, following traditional monolingual KBQA methods \cite{saxena2020improving,he2021improving}. 
We then convert the subgraph into a question-specific passage with KB-to-text generation models, incorporating the KB triples with contextual expressions. 
Given the converted cross-lingual question-passage pairs, we adopt MPLMs to rank answer candidates in the passages. 
As a general framework, our approach can be easily applied to different languages or KBs without specialized modifications.

We empirically investigate the effectiveness of our method on two xKBQA datasets, QALD-M~\cite{ngomo20189th} and WebQSP-zh.
QALD-M is a collection of a few hundred questions in 11 non-English languages, from a series of xKBQA evaluation campaigns.
Considering its small size, we also construct a new dataset WebQSP-zh with 4,737 Chinese questions translated from WebQSP~\cite{yih2016value} by native speakers. 
WebQSP-zh is much larger in size and involves more natural expressions as the annotators take into account commonsense knowledge and realistic vocabulary choices during manual translation. 

Experimental results demonstrate that our method outperforms a variety of English-as-pivot baselines based on monolingual KBQA models, reaching 74.37\% hits@1 on WebQSP-zh. 
Moreover, our method achieves strong few-shot and zero-shot performance. 
Using only 10\% of the training data, our method performs comparably to several competitive English-as-pivot baselines trained with full training data. 
For the zero-shot evaluation on QALD-M, our method achieves 51.20\% hits@1 on average across 11 languages.

Our main contributions are summarized as:
\vspace{-\topsep}
\begin{itemize}
    \item 
    We formulate xKBQA as answering questions by reading passages converted from KB subgraphs, bridging the gap between KB schemas and natural language expressions. Existing high-quality xMRC resources are further utilized to alleviate the data scarcity issue.
    \vspace{-\topsep}
    \item 
    We collect a large xKBQA dataset with native expressions in Chinese, i.e., WebQSP-zh. It, along with its original version, i.e., WebQSP, can be used for analyzing the gap between monolingual and cross-lingual KBQA.
    \vspace{-\topsep}
    \item We conduct extensive experiments on two datasets with 12 languages. Our method outperforms various baselines and achieves strong few-shot and zero-shot performance.
\end{itemize}

\begin{figure*}[ht]
\centering
\includegraphics[scale=0.23]{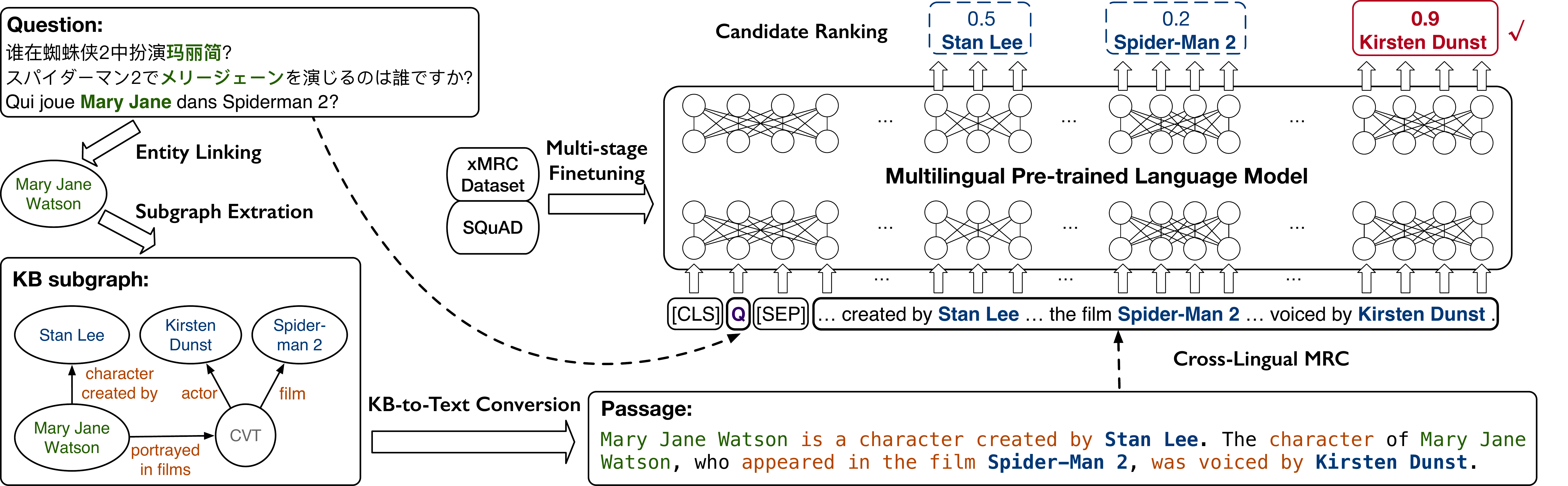}
\caption{An illustration of our method. 
Given a large-scale KB in a rich-resource language such as English, 
to answer a question in a language with relatively fewer resources, we first extract a subgraph from KB according to entity linking results. 
We then convert the subgraph into a question-specific passage in natural language with KB-to-text generation models, complementing the KB triples with contextual expressions. The question and converted passage are fed into a PLM-based xMRC model, which ranks all candidate answer spans to obtain the final answer.
}
\label{fig:illustration}
\end{figure*}

\section{Related Works}
\paragraph{KBQA} 
Recent efforts in KBQA generally fall into two main paradigms, either the information extraction style~\cite{miller-etal-2016-key,sun2018open,xu-etal-2019-enhancing,saxena2020improving,he2021improving,shi-etal-2021-transfernet}
or the semantic parsing style~\cite{yih-etal-2015-semantic,lan-jiang-2020-query,ye-etal-2022-rng,gu-su-2022-arcaneqa}. 
The former retrieves a set of candidate answers from KB, which are then compared with the questions in a condensed feature space. 
The latter manages to distill the symbolic representations or structured queries from the questions. 

\paragraph{xKBQA} 
Both styles of KBQA methods can be applied to xKBQA. 
Previous xKBQA efforts generally fall in the semantic parsing style.
They rely on online translation tools \cite{hakimov2017amuse} or embedding-based word-to-word translation \cite{zhou2021improving} to obtain synthetic training data. 
In contrast, the information extraction based xKBQA approach is less explored. An advantage of this style of xKBQA methods is that it requires no annotation of structured queries, which is expensive to obtain for non-English languages. 
In this paper, we attempt to explore xKBQA approaches of the information extraction style with less reliance on machine translation tools and investigate their performance in the few-shot and zero-shot settings.

\paragraph{xMRC} 
xMRC is a cross-lingual QA task receiving extensive attention recently, with considerable progress in datasets and models. There has been a stream of high-quality datasets in a wide range of languages, including MLQA \cite{lewis2020mlqa}, MKQA \cite{longpre-etal-2021-mkqa}, XQuAD \cite{artetxe2020cross} and TyDi QA \cite{Clark2020TyDiQA}. 
Several works for xMRC adopt machine translation tools\cite{Asai2018MultilingualER,Cui2019CrossLingualMR,Lee2019LearningWL}
or question generation systems \cite{riabi-etal-2021-synthetic} to obtain more cross-lingual training data, while other works attempt to learn better cross-lingual mapping with MPLMs \cite{Yuan2020EnhancingAB,wu-etal-2022-learning}. 

\paragraph{KB-to-text in QA} 
To benefit xKBQA with the progress in xMRC, we propose to convert the xKBQA task into reading comprehension. 
Previous works in other QA tasks attempt to convert KB triples into texts by simple concatenating heuristics \cite{oguz2021unik} or by manually-designed rules \cite{bian2021benchmarking}. 
\citet{ma-etal-2022-open} resort to PLM-based generation models and argue that data-to-text can serve as a universal interface for open domain QA.
To the best of our knowledge, our work is the first to introduce data-to-text methods into KBQA and cross-lingual QA. 
Compared with \citet{ma-etal-2022-open}, 
we further address the real-world problems of complex KB structures, cross-lingual semantic gap, and data scarcity when applying data-to-text to xKBQA. 

\section{Methodology}
We propose a novel approach to tackle xKBQA as reading comprehension.
As illustrated in Figure \ref{fig:illustration}, we first convert KB triples into sentences using generation models and obtain question-specific passages for reading comprehension.  
We then adopt MPLMs finetuned on xMRC datasets to answer cross-lingual questions according to the converted passages. 

\subsection{Task Formulation}
In xKBQA, given a knowledge base $G$ in language $A$ and a question $q$ in another language $B$, the model is expected to answer $q$ by entities or literal values in $G$.
In practice, $A$ is often a rich-resource language such as English, and $B$ is a language with relatively fewer resources. 
A knowledge base $G$ consists of a set of knowledge triples.
In a triple $(h, r, t)$, $h \in E$ is a head entity, $t \in E \cup L $ is a tail entity or a literal value, and $r \in R$ is the relation/predicate between $h$ and $t$, where $E$ denotes the set of all entities, $L$ denotes the set of all literal values, and $R$ denotes the set of all relations.

\subsection{KB-to-Text Conversion}

In a typical monolingual KBQA framework, one first identifies the topic entity in the question and links it to the given KB. This can be achieved by surface-level matching \cite{sun2018open} or supervised entity linkers \cite{yang2015s}. 
In the cross-lingual setting, one can directly adopt multilingual entity linkers such as mGENRE~\cite{de-cao-etal-2022-multilingual} or translate questions and KB entities into the same language for monolingual linking.  

After entity linking, a KB subgraph is constructed by the neighbors within several hops around the topic entities. 
Based on the given question, all candidates in the subgraph are ranked to arrive at the final answers. 
To successfully identify from the subgraph the KB predicates leading to the answer, the KBQA models are expected to learn a mapping between KB predicates and natural language expressions in the questions.
In addition to the language gap as in most cross-lingual tasks, the models have to deal with the difference in expression styles used in the KB schemas and questions.

To narrow down the gap of mapping, we propose to convert KB subgraphs to natural language passages, formulating xKBQA as an xMRC task, so that we can benefit from recent advances in xMRC.
Converting KB subgraphs into natural sentences brings plausible context for candidate KB answers, facilitating the matching between questions and answers.
Furthermore, with the natural language expressions of the KB subgraphs, current xMRC models can be directly adopted to solve the questions. 
We believe that xMRC models could benefit the xKBQA task for their strong capabilities of mapping between cross-lingual expressions.
Even without annotated xKBQA data, they are able to answer a portion of xKBQA questions, utilizing their prior knowledge of the cross-lingual mapping learned from pre-training and fine-tuning on xMRC datasets.

\begin{figure}[t]
\centering
\includegraphics[scale=0.3]{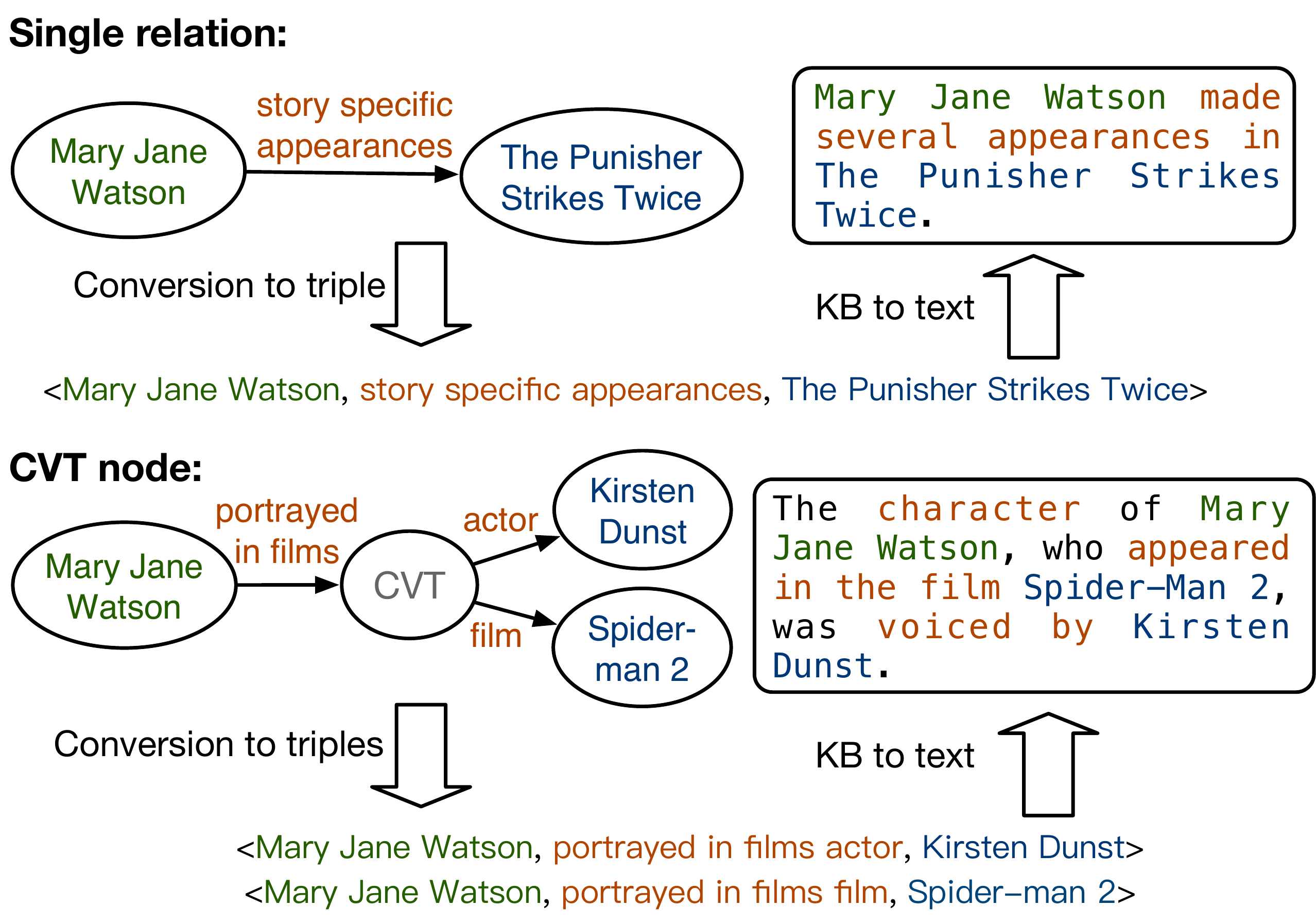}
\caption{Examples of KB-to-text conversion for a single relation (upper) and a complex event-like fact, such as CVT nodes in Freebase (lower).}
\label{fig:kb2text}
\end{figure}

To convert KB subgraphs into readable passages, we utilize PLM-based KB-to-text models, such as JointGT~\cite{chen2021jointly}.
A KB-to-text model converts a structured KB subgraph to natural language texts, complementing the given entities and relations with potential contextual expressions. 
Compared with simply concatenating the head entity, relation and tail entity of a triple, a KB-to-text model can generate more natural and coherent sentences. 
It also alleviates the onerous manual design of conversion rules.
Moreover, the KB-to-text model can handle not only single-relation triples but also more complex KB structures, such as CVT nodes, which is a complex node type in Freebase referring to an event with multiple fields. 
Figure \ref{fig:kb2text} shows examples of KB-to-text conversion for a single-relation triple and a CVT node.

After conversion, we identify the candidate answer spans from the pieces of text with fuzzy string matching tools. 
To form a passage, we concatenate the pieces of text, sorted by their semantic similarities to the questions.\footnote{Previous work shows that PLM-based MRC models are not sensitive to the order of sentences in the passage \cite{sugawara2020assessing}. We do not observe significant performance change after we shuffle the sentence order in the passage, which conforms to the finding by \citet{sugawara2020assessing}.}
We observe that the subgraphs around a topic entity can be very large, especially for the \textit{hub} entities like \textit{the USA}. 
Consequently, the converted passages can be very long, even up to 20k words in length. 
Current xMRC models struggle with such long passages. 
To shorten the converted passages, we fix the maximum length of the passage and discard the remaining redundant sentences.

\subsection{Cross-Lingual Reading Comprehension}
MPLMs are widely adopted in xMRC for their strong capabilities of understanding cross-lingual texts. 
They can encode different languages in a unified semantic space, relieving the reliance on translation tools.
We thus use MPLMs to solve the xMRC instances converted from xKBQA.

Specifically, we concatenate the question and the converted passage as the input to the MPLMs and predict the boundary of the answer span.
In the KB-to-text step, we have identified the corresponding span in the passage for each candidate KB entity or literal value. 
Thus, during inference, we only need to rank the candidate answer spans.
The corresponding KB entity or value for the top-ranked candidate span is selected as the final answer.

To address the data scarcity in xKBQA, we further propose to finetune the models on MRC data in multiple stages before on xKBQA data.
Compared to KBQA, it is easier to acquire annotated MRC data for its straightforward annotation process without the requirement of background knowledge in KBs.
Apart from large-scale English MRC datasets such as SQuAD~\cite{Rajpurkar2016SQuAD1Q}, there are a series of high-quality xMRC datasets, including MLQA, MKQA and XQuAD, covering a wide range of non-English languages such as Russian, Hindi, and Dutch. 
In the first stage, we use large-scale English MRC datasets, e.g., SQuAD, to help MPLMs learn the language-agnostic ability to find answers from the passages. 
In the second stage, we finetune the models on high-quality xMRC datasets in the target language, strengthening the reading comprehension ability for the target language. 
In this way, the two-stage finetuning before training on xKBQA data benefits models with the rich resources in MRC and mitigate the data scarcity problem in xKBQA.

\section{Experimental Setup}

\begin{table}
\small
\centering
\begin{tabular}{p{0.94\columnwidth}}
\toprule
\textbf{WebQSP-zh:}
\begin{CJK*}{UTF8}{gbsn}\small安娜肯德里克出演过什么？\end{CJK*}/ \textit{What did Anna Kendrick star in?}\\
\textbf{WebQSP-MT:}
\begin{CJK*}{UTF8}{gbsn}\small安娜肯德里克在干什么？\end{CJK*}/ \textit{What is Anna Kendrick doing?}  \\
\textbf{WebQSP:} What has Anna Kendrick been in? \\
\textbf{Freebase Predicate:} film.actor.film~~film.performance.film \\
\midrule
\textbf{WebQSP-zh:} \begin{CJK*}{UTF8}{gbsn}\small1945年前苏联的领导人是谁？\end{CJK*}/ \textit{Who was the leader of the former Soviet Union in 1945?} \\ 
\textbf{WebQSP-MT: } \begin{CJK*}{UTF8}{gbsn}\small1945年苏联的领导人是谁？\end{CJK*}/ \textit{Who was the leader of the Soviet Union in 1945?} \\
\textbf{WebQSP:} Who was the leader of the Soviet Union in 1945? \\
\textbf{Freebase Predicate:} \\
government.governmental\_jurisdiction.governing\_officials\\
government.government\_position\_held.office\_holder \\
\bottomrule
\end{tabular}
\caption{Examples from WebQSP-zh and their corresponding questions in WebQSP. 
WebQSP-MT is the Chinese translation of WebQSP by Baidu Translate, a machine translation tool. The italic English texts are the literal meaning of the Chinese questions. 
}
\label{tab:webqspzh}
\end{table}

\subsection{Datasets} 
We evaluate our method on two datasets, \textbf{QALD-M}, a small evaluation dataset in 11 languages, and \textbf{WebQSP-zh}, a new dataset with a larger size and more realistic expressions. 

\paragraph{QALD-M} QALD-M is a series of evaluation campaigns on question answering over linked data. 
We use the version provided by \citet{zhou2021improving} and filter the out-of-scope ones.
It consists of testing questions for 11 non-English languages (fa, de, ro, it, ru, fr, nl, es, hi, pt, pt\_BR) over DBPedia. 
The numbers of used questions for each language range from 66 to 363. We use QALD-M mainly for zero-shot evaluation. 
See Appendix \ref{app:qald-m} for more details.

\paragraph{WebQSP-zh}
Considering that the size of QALD-M is small and its multilingual questions are mostly literal translations without language-dependent paraphrasing, we collect a new xKBQA dataset WebQSP-zh, with 3,098 questions for training and 1,639 questions for test. 

To collect WebQSP-zh, we employ two Chinese native speakers proficient in English to manually translate all the questions in WebQSP \cite{yih2016value}, a widely-used English KBQA dataset, together with another annotator responsible for checking translation quality.
To provide a more realistic benchmark for cross-lingual evaluation, the annotators are instructed to pay much attention to commonsense knowledge and natural vocabulary choices during translation. 
For example, in the upper example of Table~\ref{tab:webqspzh}, the phrase \textit{be in} in the WebQSP question has multiple translations in Chinese. Based on the commonsense knowledge that \textit{Anna Kendrick} is an actress, it is translated as \begin{CJK*}{UTF8}{gbsn}\small出演\end{CJK*}/\textit{star in} instead of its literal meaning \begin{CJK*}{UTF8}{gbsn}\small在做\end{CJK*}/\textit{be doing}. 
In the lower example of Table~\ref{tab:webqspzh}, the annotator chooses the Chinese word \begin{CJK*}{UTF8}{gbsn}\small前苏联\end{CJK*}/\textit{former Soviet Union} for translation instead of \begin{CJK*}{UTF8}{gbsn}\small苏联\end{CJK*}/\textit{Soviet Union} because the former is more often used by native Chinese speakers. 
See Appendix \ref{app:weqsp-zh} for more statistics, annotation details, and examples.

\subsection{Baselines} 
\paragraph{Supervised}
A widely-adopted baseline method in cross-lingual QA tasks is translating data in non-English languages into English with machine translation tools and utilizing mono-lingual models~\cite{Asai2018MultilingualER,Cui2019CrossLingualMR}, which we call \textbf{English-as-pivot}. 
For supervised experiments on WebQSP-zh, we select several competitive monolingual KBQA models for English-as-pivot evaluation.
For information extraction style, we select \textbf{EmbedKGQA} \cite{saxena2020improving}, \textbf{GraftNet} \cite{sun2018open}, \textbf{NSM} (with its teacher-student variant, \citealp{he2021improving}), all of which require no annotation of structured KB queries, as our method does.
For semantic parsing style, we select \textbf{QGG}~\cite{lan-jiang-2020-query}.
\footnote{We did not include the recent semantic-parsing-style models based on Seq2Seq generation, including RnG-KBQA~\cite{ye-etal-2022-rng} and ArcaneQA~\cite{gu-su-2022-arcaneqa}, both of which outperform QGG by 1.6\% F1 on WebQSP. However, setting up an environment for them requires up to 300G memory, far exceeding our computational budgets.
So we think that OGG is a suitable baseline that strikes a good balance between performance and computational resources.}

We also provide a \textbf{Closed-book QA} baseline~\cite{roberts-etal-2020-much} with generation-based MPLMs, e.g., mT5~\cite{xue-etal-2021-mt5}. 
We feed the question directly into the model and expect it to output the answer based on its knowledge learned in pre-training.
This method requires no external knowledge, such as KBs, and can coarsely evaluate how much parametric knowledge an MPLM may have. 

\paragraph{Zero-shot}
Since the above supervised baselines are unable to answer any questions without training data, we further implement two baselines inspired from \citet{zhou2021improving} for zero-shot evaluation.
One is \textbf{Multilingual Semantic Matching}, which measures the similarity between questions and inferential chains with an MPLM finetuned on LC-QuAD \cite{trivedi2017lc}, an English KBQA dataset. 
The other, based on the previous baseline, uses \textbf{Bilingual Lexicon Induction} (BLI, \citealp{lample2018word}) to obtain word-to-word translation in the target languages as data augmentation.

\subsection{Metrics}
Following previous works \cite{saxena2020improving,he2021improving}, we use hits@1 as the evaluation metric. 
It is the ratio of questions whose top~1 predicted answer is in the set of golden answers.

\subsection{Implementation Details}
Following previous works~\cite{sun2018open,saxena2020improving,he2021improving}, we use the golden topic entities for a fair comparison with the baselines. 
We also discuss the effects of entity linking in Section \ref{sec:el}. 
For KB-to-text generation, we use JointGT \cite{chen2021jointly} finetuned on WebNLG \cite{gardent2017creating}, a KB-to-text dataset.
We use TheFuzz\footnote{\url{https://github.com/seatgeek/thefuzz}} to identify candidate answer spans.
We fix the maximum passage length to 750 words and discard the sentences with lower semantic similarity to the questions, measured by the multilingual model of SentenceTransformers \cite{Reimers2020MakingMS}. 
For xMRC, we experiment with mBERT and XLM-R. 
Before finetuning on the xMRC instances converted from xKBQA datasets, we first finetune models on SQuAD~1.1, and then on three xMRC datasets, MLQA, MKQA and XQuAD. 
We do not search hyperparameters for the xMRC models and adopt the default configuration used by SQuAD.
For English-as-pivot baselines, we use Baidu Translate API\footnote{\url{https://fanyi.baidu.com/}} to obtain English translations. 
See Appendix \ref{app:implementation} for more  details.

\section{Results and Analyses}
\subsection{Supervised Setting}

\begin{table}
\centering
\small
\begin{tabular}{lll}
\toprule
\textbf{Model}  & \textbf{WebQSP} & \textbf{WebQSP-zh}  \\
\midrule
\multicolumn{1}{l}{\textit{English-as-pivot}} \\
EmbedKGQA~\citeyearpar{saxena2020improving} & 66.18 & 63.15 (-3.03) \\
GraftNet~\citeyearpar{sun2018open} & 67.79 & 65.61 (-2.18)\\
NSM~\citeyearpar{he2021improving} & 68.70 & 67.30 (-1.40) \\
NSM-student~\citeyearpar{he2021improving} & 74.30 & 72.54 (-1.76)\\

QGG~\citeyearpar{lan-jiang-2020-query} & 73.70 & 72.36 (-1.34) \\

\midrule
\multicolumn{1}{l}{\textit{Closed-book QA}} \\
mT5-base & & 7.02 \\
mT5-large & & 12.87 \\
\midrule
\multicolumn{1}{l}{\textit{xKBQA-as-MRC (Ours)}} \\
mBERT-base & & 70.53 \\ 
XLM-R-base & & 69.92\\
XLM-R-large & & \textbf{74.37} \\

\bottomrule
\end{tabular}
\caption{Hits@1 (\%) of baselines and our method on the test set of WebQSP-zh using the full training data. The ``WebQSP'' column shows the model performance on the test set of WebQSP after training on the original English WebQSP data. The numbers in the brackets denote the performance drop of English-as-pivot models compared to their corresponding English KBQA models on WebQSP. 
All models except GraftNet use golden topic entities.
}
\label{tab:main_results}
\end{table}

\begin{table*}[t]
\small
\centering
\begin{tabular}{lccccccccccc|c}
\toprule
\textbf{Model} & \textbf{fa} & \textbf{de} & \textbf{ro} & \textbf{it} & \textbf{ru}  & \textbf{fr} & \textbf{nl} & \textbf{es} & \textbf{hi} & \textbf{pt} & \textbf{pt\_BR} & \textbf{Avg.} \\
\midrule
\multicolumn{13}{l}{\textit{Multilingual Semantic Matching}} \\
LC-QuAD & 43.41 & 44.90 & 48.55 & 47.93 & 36.84 & 47.38 & 43.93 & 46.53 & 41.60 & 37.43 & 48.48 & 44.27 \\
+ Sing. BLI & 46.41 & 50.41 & 50.87 & 51.24 & 40.35 & 48.76 & 48.55 & 49.42 & 34.73 & 40.35 & 54.54 & 46.88 \\
+ All BLI & 46.41 & 49.31 & 50.58 & 49.04 & 41.52 & 49.59 & 47.40 & 48.55 & 41.98 & 40.94 & 51.51 & 46.98\\
\midrule
\multicolumn{13}{l}{\textit{xKBQA-as-MRC (Ours)}} \\
SQuAD & 39.22  & 48.21 & 44.48 & 45.45 & 33.33 & 45.17 & 48.27 & 47.11 & 43.89 & 35.67 & 51.51 & 43.85    \\

+ Sing. xMRC &  39.22 & 52.07 & \textbf{52.91} & \textbf{56.20} & \textbf{45.61} & 51.24 & 52.02 & \textbf{54.62} & \textbf{50.76} & \textbf{42.69} & 59.09 & 50.59     \\

+ All xMRC & \textbf{48.50} & \textbf{55.10} & 52.03 & 54.27 & 44.44 & \textbf{53.44} & \textbf{52.89} & 53.47 &  46.95 & 41.52 & \textbf{60.61} & \textbf{51.20} \\

\bottomrule
\end{tabular}
\caption{Hits@1 (\%) of the baseline and our method with XLM-R-large on QALD-M under the zero-shot setting. ``LC-QuAD'' and ``SQuAD'' means using LC-QuAD and SQuAD for finetuning, respectively. ``BLI'' and ``xMRC'' means using BLI translation and xMRC datasets for finetuning, respectively. `` Sing.'' means using the data in the target language only while ``All'' means combining the data in all the languages. We do not find available xMRC datasets for Persian (fa), so the performance  of ``+ Sing. xMRC'' on Persian is the same as that of ``SQuAD''. }
\label{tab:QALD-M}
\end{table*}

As shown in Table \ref{tab:main_results}, we first compare our method with English-as-pivot baselines using full training data of WebQSP-zh. 
These baselines can benefit from the development of monolingual KBQA models and achieve over 63\% hits@1 on WebQSP-zh. 
Suppose we have perfect translation results, the English-as-pivot baselines on the WebQSP-zh should reach the performance of monolingual models on the original English WebQSP. 
However, the English-as-pivot baselines on WebQSP-zh drop 1.4-3.0\% hits@1 compared to their monolingual performance  on the original WebQSP. 
This is because the English-as-pivot baselines are highly dependent on machine translation tools, whose outputs may contain unnatural expressions or even errors.

As for the closed-book QA baselines, mT5-large correctly outputs the answers in English for even 12.9\% of the WebQSP-zh questions, without resorting to any external knowledge. 
This proves that MPLMs have learned a large amount of factual knowledge and strong cross-lingual capabilities, which can be properly utilized for xKBQA, as our method does.

All our models reach over 69\% hits@1 on WebQSP-zh. 
Our two base-size models outperform EmbedKGQA by approximately 6\% hits@1, an English-as-pivot baseline that utilizes RoBERTa-base and the KB embedding ComplEx~\cite{trouillon2016complex}. 
Our model with XLM-R-large outperforms all baselines, achieving 74.37\% hits@1 thanks to the strong cross-lingual capability from MPLMs and rich resources in xMRC. 
Moreover, these results demonstrate another merit of our approach that it can directly answer non-English questions over KBs in English, reducing the reliance on machine translation systems.
Although NSM-student, which does not use PLMs itself, performs better than our two base-size models, 
the parameters and computational complexity introduced by the translation system are much heavier than the MPLM used in our method.
Furthermore, our approach demonstrates its advantage with fewer or even no training data, as we will discuss next.

\subsection{Few-Shot and Zero-Shot Settings} \label{sec:few-shot}

\begin{figure}[t]
\centering
\includegraphics[scale=0.4]{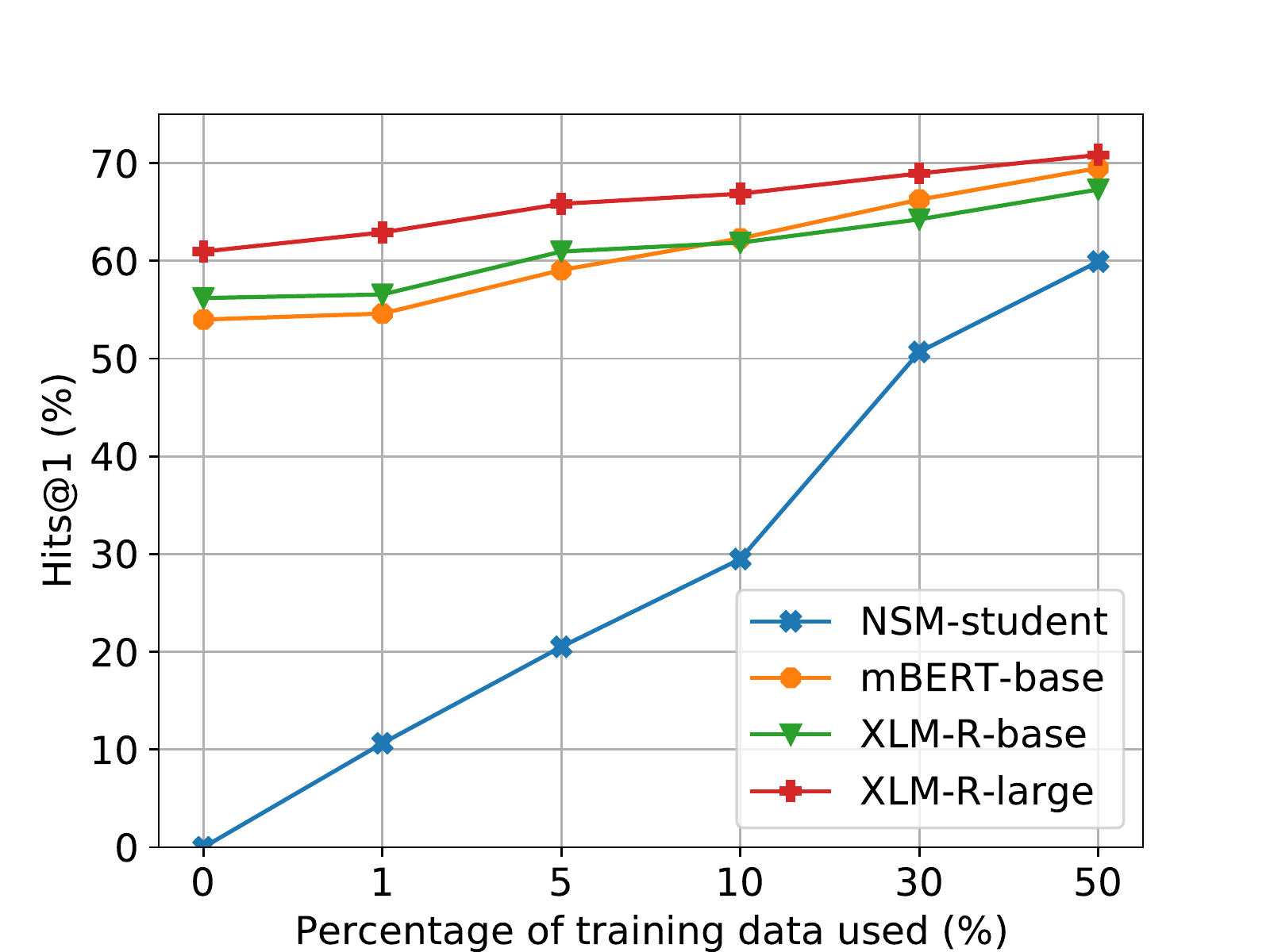}
\caption{Few-shot and zero-shot performance of our method and NSM-student on the test set of WebQSP-zh.}
\label{fig:few-shot}
\end{figure}

Consider the high cost of annotating high-quality xKBQA data,
we investigate the capabilities of our method under few-shot and zero-shot settings.

Figure \ref{fig:few-shot} shows the performance of our method and NSM-student on WebQSP-zh under few-shot and zero-shot settings. 
For NSM-student, its performance drops drastically with the decrease in training data.
and it is totally incapable of zero-shot xKBQA.
By contrast, when trained with half of the training data, our method
still performs well, with less than 3\% decrease in hits@1 compared with those trained with full data. 
With only 10\% of the training data, i.e., 310 instances, our models reach over 62\% hits@1, comparable with EmbedKGQA trained with full training data. 
Even under the zero-shot setting, our method can achieve 53-61\% hits@1. 
The high performance of our method with limited training data is attributed to the KB-to-text conversion, which in turn makes it possible to benefit from the rich resources in xMRC. 
The MPLMs for xMRC have learned to encode different languages in the same semantic space during pre-training.
After finetuning on xMRC datasets, the models can learn the ability to seek information from passages in a different language. 
By combining the prior knowledge of cross-lingual mapping and reading comprehension abilities, our models can successfully answer a large portion of the xMRC-like questions converted from xKBQA.

To demonstrate that our method can generalize to different languages without specialized modifications, we test our approach on QALD-M in 11 typologically-diverse languages under the zero-shot setting.  
We evaluate the model on QALD-M after finetuning (1) on SQuAD only, (2) on SQuAD and xMRC datasets of a single language, and (3) on SQuAD and xMRC datasets of all the languages.
As shown in Table \ref{tab:QALD-M}, after finetuning XLM-R-large with SQuAD, our models achieve 43.9\% hits@1 on average across 11 non-English languages, 
demonstrating our method's strong generalization ability from English MRC datasets. 
After further finetuning on xMRC datasets for each language, we observe a 6.7\% hits@1 boost in the average performance, showing the benefit of xMRC datasets in the  absence of xKBQA data. 
If we combine the xMRC of all the languages for finetuning, the average hits@1 further increases slightly by 0.6\%, probably due to the potential complementary effects between data in different languages.
Compared with the semantic matching baseline finetuned with LC-QuAD and BLI-based translations, our best model outperforms it by 4.2\% hits@1 on average. This is because the KB-to-text process of our method provides richer context than single inferential chains and the xMRC data are of higher quality than the BLI-based word-to-word translation. 

\subsection{Ablation Study}
To evaluate the effectiveness of the designs in our approach, we conduct experiments in several ablated settings on WebQSP-zh with full xKBQA training data.
We additionally conduct an ablation study with only 10\% of the training to investigate what is behind the promising few-shot performance.
The results are shown in Table \ref{tab:ablation}.

With full training data, after we replace the PLM-based KB-to-text model with the simple heuristic of concatenating the head, predicate, and tail (w/o KB to text), the performance drops by 2.13\% hits@1.
Although the xMRC models can to some extent learn the mapping between questions and sentences converted by heuristics, the coherence and readability of KB-to-text generation results contribute to the final performance. 
Skipping the finetuning on either SQuAD (w/o SQuAD) or xMRC datasets (w/o xMRC data) leads to a performance drop, showing the importance of high-quality data augmentation in absence of large-scale xKBQA data.

In the setting with 10\% of the training data, both KB-to-text generation and finetuning on the MRC data contribute to the high few-shot performance, similar to the full training data setting. 
We observe a drastic drop of 12.81\% hits@1 if the model is not finetuned on any MRC data (w/o xMRC data, SQuAD). 
This indicates that MRC data, no matter monolingual or cross-lingual, can greatly relieve the problem of data scarcity in xKBQA.

\begin{table}
\centering
\small
\begin{tabular}{lll}
\toprule
\textbf{Model}  & \textbf{100\%} & \textbf{10\%} \\
\midrule
XLM-R-large (Ours) &  74.37  & 67.60 \\
- w/o KB to text & 72.24 \scriptsize{(-2.13)} & 65.58 \scriptsize{(-2.02)}\\
- w/o xMRC data  & 71.81 \scriptsize{(-2.56)} & 65.53 \scriptsize{(-2.07)} \\
- w/o SQuAD      & 71.02 \scriptsize{(-3.35)} & 65.10 \scriptsize{(-2.50)}\\
- w/o xMRC data, SQuAD & 66.69 \scriptsize{(-7.68)} & 54.79  \scriptsize{(-12.81)}\\
\bottomrule
\end{tabular}
\caption{Ablation study of our method  with XLM-R-large on WebQSP-zh, using 100\% or 10\% of the training data (Hits@1 in percent). }
\label{tab:ablation}
\end{table}

\begin{table*}[ht]
\small
\centering
\begin{tabular}{p{0.25\columnwidth}p{0.95\columnwidth}p{0.6\columnwidth}r}
\toprule
\textbf{Source}  & \textbf{Example} & \textbf{Explanation} & \textbf{\%} \\
\midrule

Answer

Annotation
& 
\textbf{Question:} \begin{CJK*}{UTF8}{gbsn}\small沃尔玛经营什么产业？\end{CJK*}/ \textit{What industry does Walmart operate in?} 

\textbf{Passage:} ... The industry of Walmart is \underline{Retail-Store}, \underline{Variety Stores} and \underline{Department Stores}. ...

\textbf{Answer:} Variety Stores

\textbf{Prediction:} Retail-Store
& The annotated answers in the original WebQSP dataset are  incomplete or incorrect. In the left case, the annotated answer set fails to include two correct answers, \textit{Retail-Store} and \textit{Department Stores}.
& 34 \\
\midrule

KB-to-text

Generation
& \textbf{Question:} \begin{CJK*}{UTF8}{gbsn}\small凯南·鲁兹在灯红酒绿杀人夜中扮演谁？\end{CJK*}/ \textit{Who does Kellan Lutz play in Prom Night?} 

\textbf{Passage:} ... Kellan Lutz, a character in the film ``Prom Night'', played with \underline{Rick Leland}. ... Kellan Lutz, a character in Twilight, played the role of \underline{Emmett Cullen}. ...

\textbf{Answer:} Rick Leland

\textbf{Prediction:} Emmett Cullen
& The KB-to-text model converts a KB schema
to a wrong natural language expression or omits the entities in the given triple. In the left case, the model incorrectly converts the KB schema \textit{character} to the expression \textit{play with}.
& 12 \\
\midrule

Sentence

Filtering
& \textbf{Question:} \begin{CJK*}{UTF8}{gbsn}\small爱德华多·包洛奇在他的工作中使用了什么材料？\end{CJK*}/ \textit{What Materials did Eduardo Paolozzi use in his work?} 

\textbf{Passage:} ... The art forms of Eduardo Paolozzi are \underline{Sculpture}. ... 

\textbf{Answer:} Bronze

\textbf{Prediction:} Sculpture
& The answers are missing in the passages because the model for sentence similarity calculation incorrectly filters out the sentences containing answers. In the left case, the sentence containing the answer \textit{Bronze} is mistakenly filtered out.
&  20\\
\midrule

Reading

Comprehension
& \textbf{Question:} \begin{CJK*}{UTF8}{gbsn}\small谁是杰拉尔德福特的副总裁？\end{CJK*}/ \textit{Who was the vice president of Gerald Ford?} 

\textbf{Passage:} ... \underline{David Gergen} was appointed as the White House Communications Director by President Gerald Ford . ... The vice president of Gerald Ford was \underline{Nelson Rockefeller }. ...

\textbf{Answer:} Nelson Rockefeller

\textbf{Prediction:} Staff Dick Cheney
& The xMRC model fails to select the correct answer span. In the left case, the xMRC model incorrectly maps the word \begin{CJK*}{UTF8}{gbsn}\small副总裁\end{CJK*}/\textit{vice president} to the expression \textit{White House Communications Director} in the passage. 
& 34 \\
\bottomrule
\end{tabular}
\caption{Examples, explanations and percentages of different sources of error in the 50 sampled WebQSP-zh question that XLM-R-large fails to answer. The underlined spans in passages are answer candidates.}
\label{tab:err_examples}
\end{table*}

\subsection{Error Analysis}
We sample 50 error cases in WebQSP-zh and analyze their sources of error, as shown in Table \ref{tab:err_examples}. 

34\% of the errors result from the annotation of the original WebQSP dataset, where the annotated answer sets may be incomplete or incorrect. 
Another common source of error is the MRC model, which incorrectly answers 34\% of the sampled questions. 
Among them, many are complex questions involving constraints or multiple relations. 
In the future, multi-hop MRC models can be adopted for addressing them.
Besides, there are also several error cases resulting from KB-to-text generation and sentence filtering.
We believe that our model will achieve better performance if each module in our framework is carefully optimized for the datasets.

\subsection{Effect of Entity Linking} \label{sec:el}
Entity linking (EL) is a crucial issue in KBQA, which requires linking the entity mentions in the questions to the entities in a KB. 
It becomes even more difficult in the cross-lingual setting. 
In the experiments above, we use golden entity linking results following previous works. 
To further investigate the effect of entity linking in xKBQA, we conduct pilot experiments with two EL methods. One is surface-level matching after translating the questions, and the other is mGENRE~\cite{de-cao-etal-2022-multilingual}, a cross-lingual EL tool that does not rely on machine translation tools.
On the test set of WebQSP-zh, two EL methods achieve 89.1\% and 76.8\% recall@5, respectively. 
With the results from two EL methods, our xMRC model with XLM-R-large achieves 65.9\% and 56.5\% hits@1, respectively.
The large gap compared to the results with golden topic entities indicates that more future research on cross-lingual EL is desired.

\section{Conclusion}
In this paper, we propose to formulate xKBQA as answering questions by reading passages, benefiting from the recent advance in xMRC. 
By converting KB subgraphs into passages, we narrow the gap between KB schemas and natural questions under cross-lingual settings.
The cross-lingual knowledge in MPLMs and the rich resources in xMRC alleviate the problem of data scarcity in xKBQA.
To facilitate the evaluation of xKBQA, we  collect WebQSP-zh, a new large-scale xKBQA dataset with more natural expressions. 
Extensive experiments on two datasets with 12  languages show the strong performance of our method under both supervised and zero-shot settings.

We hope that our work will inspire more efforts into xKBQA.
Several promising research directions under our framework include generating better passages for KB subgraphs, supporting more types of KBQA questions, and exploring better EL strategies for xKBQA.

\section*{Limitations}
We discuss the limitations of our work from the following four aspects:

First, our work mainly focuses on single-relation questions and CVT questions in KBQA.
We construct a new dataset WebQSP-zh based on WebQSP, which lacks complex questions with multiple constraints or relations.
Since we use a vanilla BERT-based MRC model in our framework, it has a limited capacity for solving complex KBQA questions. 
As future work, multi-hop MRC models can be adopted to address complex questions in cross-lingual KBQA. 

Second, our method is mainly designed for entity-centric QA. 
It can  handle well the answer types of KB entities or attribute values in KBQA. 
Yet its capability on other types of answers is currently unknown. 
We will consider extending our method with more diverse answer types in the future.

Third, the size of retrieved KB subgraphs is constrained by the maximum input length of PLMs. 
This could, to some extent, lower the answer coverage of the converted passages and hurt the overall performance. 
In the future, Longformer-based encoders or text summarization techniques could be explored to address this limitation.

Fourth, although using existing xMRC datasets can alleviate the data scarcity problem in xKBQA, it cannot fundamentally solve the problem of insufficient and expensive cross-lingual datasets. 
With more powerful cross-lingual PLMs, we may reduce the reliance on xMRC data.
We will explore more strategies for tackling the data scarcity problem in future work.

\section*{Acknowledgments}
This work is supported by NSFC (62161160339, 62206070).
We would like to thank the anonymous reviewers for their valuable suggestions.
Also, we would like to thank Xiao Liu and Quzhe Huang for their great help in this work.
For any correspondence, please contact Yansong Feng.

\bibliography{anthology,custom}
\bibliographystyle{acl_natbib}

\clearpage
\appendix

\section{Dataset Details}\label{app:dataset_detail}
\subsection{QALD-M}
\label{app:qald-m}
\paragraph{Statistic}
The QALD-M dataset used in our paper is based on the version released by \citet{zhou2021improving}, composed of questions from QALD-M 4 to QALD-M 9 in 11 non-English languages. We filter the yes/no questions, counting questions, and the questions whose answers cannot be found in the knowledge base. The sizes of testing questions for each language are shown in Table \ref{tab:QALD-M_statistic}, ranging from 66 to 363.

\paragraph{Knowledge Base}
For QALD-M, we use the 2016-10 version of DBPedia\footnote{\url{http://downloads.dbpedia.org/wiki-archive/downloads-2016-10.html}}. 
We discard the KB triples that are unlikely to contain answers such as page IDs and revision history, and only include information about article categories and object properties. 
For each question, we include in the subgraph the triples where the topic entity is the head entity or the tail entity, namely its one-hop neighbors.

\subsection{WebQSP-zh}
\label{app:weqsp-zh}

\paragraph{Statistics}
The WebQSP-zh dataset proposed in our paper consists of 4,737 question-answer pairs, of which 3,098 instances are for training and the remaining 1,639 instances are for testing.
The average length of questions is 12.7 characters.
The average number of answers per question is 9.8.

\paragraph{Knowledge Base} 
For WebQSP-zh, we use a preprocessed version of Freebase\footnote{\url{https://github.com/hugochan/BAMnet}}. 
Following previous works \cite{sun2018open,saxena2020improving}, we further prune it to contain only those relations that are mentioned in the dataset. 
For each question, we obtain the neighborhood graph within two hops of topic entities.

\paragraph{Annotation Details}
We recruited the annotators from a Chinese campus BBS, who are proficient in both Chinese and English.
They are instructed to translate the questions in WebQSP into Chinese and to pay attention to commonsense knowledge and natural vocabulary choice. 
They are paid 3 CNY for each question annotated, which is adequate given the participants’ demographic. 
The annotators are informed of how the data would be used.

\paragraph{More Examples}
We provide more examples from WebQSP-zh in Table \ref{tab:more_webqspzh} to show that WebQSP-zh is a more realistic benchmark for cross-lingual evaluation, incorporated with commonsense knowledge and realistic vocabulary choices.

In the first example of Table \ref{tab:more_webqspzh}, based on the knowledge that Aldi is a company, the word \textit{originate} is translated as \begin{CJK*}{UTF8}{gbsn}\small
创建\end{CJK*}/\textit{found} instead of its literal translation \begin{CJK*}{UTF8}{gbsn}\small
起源\end{CJK*}/\textit{originate}.  
In the second example of Table \ref{tab:more_webqspzh}, the annotator uses \begin{CJK*}{UTF8}{gbsn}\small
范德堡大学\end{CJK*}/\textit{Vanderbilt University} instead of \begin{CJK*}{UTF8}{gbsn}\small
范德堡\end{CJK*}/\textit{Vanderbilt} because native Chinese speakers often call Western universities by their full names and rarely drop the word \begin{CJK*}{UTF8}{gbsn}\small
大学\end{CJK*}/\textit{university}.
\begin{table}[htbp]
\centering
\small
\begin{tabular}{p{0.9\columnwidth}}
\toprule
\textbf{WebQSP-zh:} \begin{CJK*}{UTF8}{gbsn}\small阿尔迪是什么时候创建的？\end{CJK*}
/ \textit{When was Aldi founded?}\\
\textbf{WebQSP-MT:} \begin{CJK*}{UTF8}{gbsn}\small阿尔迪是什么时候起源的？\end{CJK*}
/ \textit{When did Aldi originate?}\\
\textbf{WebQSP:} When did Aldi originate? \\
\textbf{Freebase Predicate: } \\
business.employer.employees \\
business.employment\_tenure.from \\
\midrule
\textbf{WebQSP-zh:}
\begin{CJK*}{UTF8}{gbsn}\small范德堡大学的吉祥物是什么？\end{CJK*}/ \textit{What is Vanderbilt University's mascot?}\\
\textbf{WebQSP-MT:}
\begin{CJK*}{UTF8}{gbsn}\small范德堡的吉祥物是什么？\end{CJK*}/ \textit{What is Vanderbilt's mascot?}\\
\textbf{WebQSP:} What is Vanderbilt's Mascot? \\
\textbf{Freebase Predicate:} \\
education.educational\_institution.mascot \\
\bottomrule
\end{tabular}
\caption{Examples from WebQSP-zh and their corresponding questions in WebQSP. WebQSP-MT is the Chinese translation of WebQSP by the machine translation tool Baidu Translate. The italic English texts are the literal meaning of the Chinese questions.}
\label{tab:more_webqspzh}
\end{table}

\begin{table*}[htbp]
\centering
\small
\begin{tabular}{lccccccccccc}
\toprule
\textbf{Language} & fa & de & ro & it & ru  & fr & nl & es & hi\_IN & pt & pt\_BR \\
\midrule
\textbf{Size} & 334 & 363 & 344 & 363 & 171  & 363 & 346 & 346 & 262 & 171 & 66 \\
\bottomrule
\end{tabular}
\caption{The sizes of QALD-M testing questions in 11 languages used in our paper.}
\label{tab:QALD-M_statistic}
\end{table*}

\begin{table*}[htbp]
\centering
\small
\begin{tabular}{lccccccccccc}
\toprule
\textbf{Language} & zh & de & ro & it & ru  & fr & nl & es & hi\_IN & pt & pt\_BR  \\
\midrule
\textbf{Size} & 5,641 & 8,904 & 1,190 & 2,685 & 3,875 & 2,685 & 2,685 & 9,628 & 6,615 & 2,685 & 2,685 \\
\bottomrule
\end{tabular}
\caption{The number of questions in the combined xMRC datasets used in our paper.}
\label{tab:xmrc_statistic}
\end{table*}

\subsection{xMRC datasets}
We use three xMRC datasets for data augmentation. Their preprocessing details and statistics are as follows.

In terms of MLQA and XQuAD, we directly use the officially released data with English passages paired with non-English questions. 
In terms of MKQA, the passages for reading comprehension are full-length English Wikipedia articles. 
Since the Wikipedia articles are too long for PLM-based xMRC models to handle, we use the annotated non-tabular long answers as passages, which are generally a few hundred words long.

For each language, we combine the data from different xMRC for finetuning. 
Specifically, we use MLQA for zh, de, es, hi; MKQA for zh, de, es, fr, it, nl, pt, pt\_BR, ru; XQuAD for de, es, hi, ro, ru. 
The statistics of the combined xMRC data are shown in Table \ref{tab:xmrc_statistic}.

\section{Implementation Details} \label{app:implementation}
\subsection{KB-to-Text}
We use JointGT \cite{chen2021jointly} based on BART-base for KB-to-text generation. It is finetuned on WebNLG with the same hyperparameters in the original paper. In sentence filtering, we use the paraphrase-multilingual-mpnet-base-v2 model in SentenceTransformers for cross-lingual semantic similarity calculation.

\subsection{xMRC} 
Our implementation of xMRC models is based on the Transformers\footnote{\url{https://github.com/huggingface/transformers}}.
For the finetuning on SQuAD, we set the batch size to 12, the learning rate to 3e-5, the number of training epochs to 2, the maximum input length to 384, and the document stride to 128.
For the finetuning on xMRC datasets and the data converted from xKBQA, we use the same hyperparameters as the finetuning on SQuAD. 
The results are from single runs. 
We use an NVIDIA A40 GPU for experiments. 
An epoch on the data converted from xKBQA takes about 9 minutes.

\section{Licenses of Scientific Artifacts}
The licenses for each dataset used are as follows:
CC BY-SA 4.0 for SQuAD, Apache-2.0 License for MKQA, CC BY-SA 4.0 for XQuAD, CC-BY-SA 3.0 for MLQA, CC-BY 4.0 for WebQSP, GPL-3.0 License for LC-QuAD, and MIT License for QALD. The licenses for each model used are as follows: Apache-2.0 License for EmbedKGQA, BSD-2-Clause License for GraftNet, and Apache-2.0 License for Transformers. No license is provided by other models.

\end{document}